\pdfoutput=1
\documentclass[11pt]{article}

\usepackage[]{acl}

\usepackage{times}
\usepackage{latexsym}
\usepackage{float}

\usepackage[T1]{fontenc}
\usepackage[utf8]{inputenc}

\usepackage{microtype}

\usepackage{inconsolata}

\usepackage{subcaption}
\usepackage{graphicx}
\usepackage{multirow}
\usepackage{caption}
\usepackage{adjustbox}
\usepackage{amsmath}
\usepackage{longtable}
\usepackage{xcolor}
\usepackage{enumerate}
\usepackage{url}
\usepackage{booktabs}
\usepackage{siunitx}
\usepackage{fancyhdr}

\fancypagestyle{specialfooter}{
  \fancyhead{} 
  
  \fancyfoot[C]{To appear in \textit{Findings of the Association for Computational Linguistics: NAACL 2024}}}

\newcommand{\gptwho}{\textsf{GPT-who}}

\title{{\gptwho}: \\ An Information Density-based Machine-Generated Text Detector}
    
\author{Saranya Venkatraman \\The Pennsylvania State University\\ \texttt{saranyav@psu.edu}\\ \And Adaku Uchendu \\ MIT Lincoln Laboratory \\ \texttt{adaku.uchendu@ll.mit.edu}\\ \And Dongwon Lee \\The Pennsylvania State University\\ \texttt{dongwon@psu.edu} \\}

\begin{document}
\thispagestyle{specialfooter}

\maketitle

\begin{abstract}
    
The {\em Uniform Information Density} (UID) principle posits that humans prefer to spread information evenly during language production. We examine if this UID principle can help capture differences between Large Language Models (LLMs)-generated and human-generated texts. We propose {\gptwho}, the first psycholinguistically-inspired domain-agnostic statistical detector. This detector employs UID-based features
to model the unique statistical signature of each LLM and human author for accurate detection. 
We evaluate our method using 4 large-scale benchmark datasets and find that {\gptwho} outperforms state-of-the-art detectors (both statistical- \& non-statistical) such as GLTR, GPTZero, DetectGPT, OpenAI detector, and ZeroGPT by over $20$\% across domains.
In addition to better performance, 
it is computationally inexpensive and utilizes an interpretable representation of text articles. We find that {\gptwho} can distinguish texts generated by very sophisticated LLMs, even when the overlying text is indiscernible.
UID-based measures for all datasets and code are available at \url{https://github.com/saranya-venkatraman/gpt-who}.
\end{abstract}

% We present the largest analysis of the UID-based representations of human and machine-generated texts (over 400k articles) to demonstrate how authors distribute information differently, and in ways that enable their detection using an off-the-shelf LM without any fine-tuning. 
\section{Introduction}

The recent ubiquity of Large Language Models (LLMs) has led to more assessments of their potential risks. 
These risks include its capability to generate
misinformation \cite{zellers2019defending,uchendu2020authorship}, 
memorized content \cite{carlini2021extracting}, 
plagiarized content \cite{lee2023language}, 
toxic speech \cite{deshpande2023toxicity}, 
and hallucinated content \cite{ji2023survey,
shevlane2023model}. To mitigate these issues, researchers have proposed automatic and human-based approaches to distinguish LLM-generated texts (i.e., machine-generated) from human-written texts \cite{zellers2019defending,pu2022deepfake,uchendu2023attribution,mitchell2023detectgpt}. 

% Such automatic detectors leverage supervised and unsupervised learning approaches to achieve accurate detection of machine-generated texts. 

\begin{figure}[H]
\centering
    \includegraphics[width=\columnwidth]{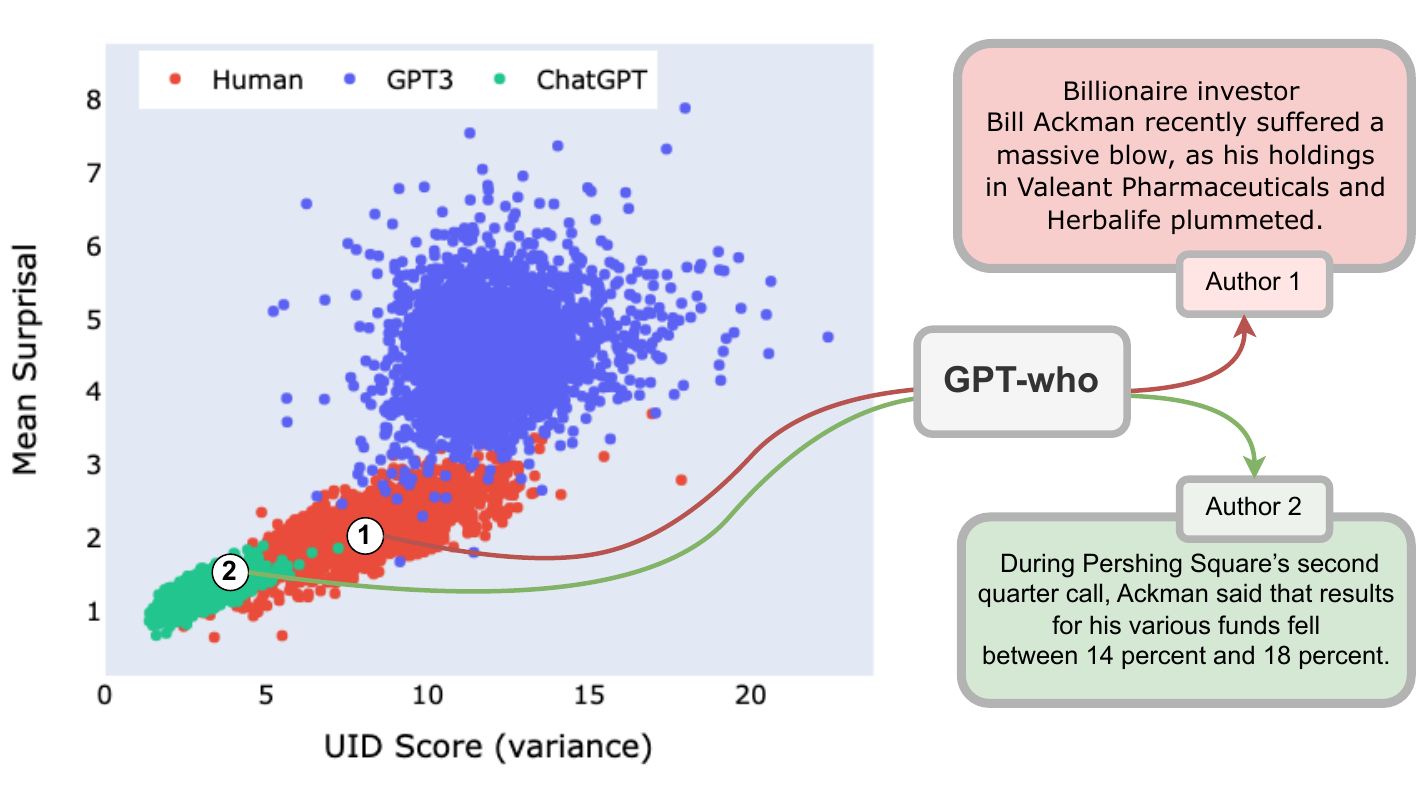}
  \caption{{\gptwho} leverages psycholinguistically motivated representations that capture authors' information signatures distinctly, even when the corresponding text is indiscernible.}
  \label{fig:teaser}
\end{figure}

Automatically detecting machine-generated texts occurs in two settings- \textit{Turing Test} (TT) which is the binary detection of human vs. machine; and \textit{Authorship Attribution} (AA) which is the multi-class detection of human vs. several machines (e.g., GPT-3.5 vs. LLaMA vs. Falcon) \cite{uchendu2021turingbench}.
While the TT problem is more rigorously studied, due to 
the wide usage of different LLMs, in the future, it will be imperative to build models for the AA tasks to determine which LLMs are more likely to be misused. This knowledge will be needed by policymakers when they inevitably institute laws to guard the usage of LLMs. 

To that end, 
we propose {\gptwho}, the first psycholinguistically-inspired supervised domain-agnostic task-independent multi-class statistical-based detector. 
{\gptwho} calculates interpretable Uniform Information Density (UID) based features from the statistical distribution of a piece of text and automatically learns the threshold (using Logistic Regression) between different authors. Such features are calculated using the surprisals of tokens in a text, for example, the variance of surprisals, the mean, and so on (elaborated in Section \ref{sec:features}). 
% However, these techniques have limitations: 
% (1) supervised learning approaches utilize deep learning or feature-based techniques which are computationally heavy and require decently sized datasets to perform well;
% (2) unsupervised techniques are all binary statistical-based approaches that require heuristic techniques to define thresholds used to distinguish these two main authors (LLM vs. Human). 
To showcase the detection capabilities of {\gptwho}, we use 4 large LLM benchmark datasets: 
TuringBench \cite{uchendu2021turingbench},
GPABenchmark \cite{liu2023check}, 
ArguGPT \cite{liu2023argugpt}, and
Deepfake Text in-the-wild \cite{li2023deepfake}. 
We find that {\gptwho} outperforms state-of-the-art statistical detectors and is at par with task and domain-specific fine-tuned LMs. This performative gain is consistent across benchmark datasets, types of LLMs, writing tasks, and domains. 

In addition to improved detection performance, {\gptwho} is computationally inexpensive as it eliminates the need for any LLM fine-tuning. It utilizes a freely available off-the-shelf LM to compute token probabilities, followed by logistic regression using a small set of carefully crafted and theoretically motivated UID features. {\gptwho} also provides a means to interpret and understand its prediction behaviors due to the rich feature space it learns from. UID-based features enable observable distinctions in the surprisal patterns of texts, which help in understanding {\gptwho}'s decision-making on authorship (Figure \ref{fig:teaser}). 

We also analyze the UID feature distributions of different LLMs and human-generated texts for the 4 datasets and find that humans distribute information more unevenly and diversely than models. In addition, UID features are reflective of differences in LLM architectures or families such that models that share architectures have similar UID distributions within but not outside their category. 
We find that UID-based features are a consistent predictor of authorship. Even when there are no glaring differences between uniform and non-uniform text, the differences in UID distributions are easily detectable and a powerful predictor of authorship, since they capture patterns that go beyond the lexical, semantic, or syntactic properties of text. Our work indicates that psycholinguistically-inspired tools can hold their ground in the age of LLMs and a simpler theoretically-motivated approach can outperform complex and expensive uninterpretable black-box approaches for machine text detection.

\section{Related Work}
\subsection{Uniform Information Density (UID)} \label{sec:UID}

Shannon's Information Theory states that information exchange is optimized when information travels across the (noisy) channel at a uniform rate \cite{shannon1948mathematical}. For language production, this uniform rate of information content is the basis of the UID hypothesis that posits that humans prefer to spread information evenly, avoiding sharp and sudden peaks and troughs in the amount of information conveyed per linguistic unit. The information content or \textbf{``surprisal''}  of a word is inversely proportional to its probability in a given context. Less predictable words have more surprisal while highly predictable words convey lower information. 
%  For example, in the sentence
% \textit{``I enjoy listening to vinyl records"}, 
% the word \textit{``records"} is highly predictable from a semantic standpoint given prior words such as  \textit{``listening"} and  \textit{``vinyl"}. Thus, given its context, \textit{``records"} has high predictability, and thus less information content or surprisal according to Information Theory. 
% Formally, Shannon's definition of information content or \textbf{Surprisal} of a component or unit (n) is given by the inverse logarithm of its probability (p(n)) i.e.
% \begin{equation}
%     Surprisal(n) = - log\ p(n)
% \end{equation}

 UID in human language production has been studied by measuring the amount of information content per linguistic unit (number of words) or by studying any sudden changes in surprisal at the onset of a word or sentential element  \cite{,xu2016entropy,jaeger2007speakers}. A rich body of work in psycholinguistics has led to the finding that, in language production, humans try to spread information content or surprisal evenly and maintain UID through their lexical, syntactic, phonological, and semantic choices \cite{frank2008speaking,xu2018information,jaeger2010redundancy,mahowald2013info,tily2009refer}.

%  Frank and Jaeger's corpus-based study demonstrated that humans tend to use shorter elements for lower amounts of information and longer elements/sub-sequences for expressing higher amounts of information \cite{frank2008speaking}. Thus, in a way keeping the information rate close to uniform. \citet{xu2018information} extended upon this work and reported that UID is consistent at the inter and intra-sentential levels \cite{xu2016entropy, xu2018information}. \citet{jaeger2007speakers} found that speakers chose not to omit an optional function word at the onset of a less predictable phrase, but that they were more likely to omit the same word at the beginning of a more predictable phrase. \citet{jaeger2010redundancy} and \citet{mahowald2013info} consolidated previous findings that humans regulate their choices as per UID, actively distributing the information that needs to be conveyed evenly across the linguistic signal.

% \cite{tily2009refer} studied the usage of `less informative' expressions as a means of conveying meanings with higher predictability in a study that directly assessed comprehenders' ease of predicting the referent in an unfolding utterance \cite{tily2009refer} and found that speakers referred to highly predictable referents with short words. Thus, in language, humans try to spread information content or surprisal evenly and maintain UID through their lexical, syntactic, phonological, and semantic choices. 

\subsection{Machine-Generated Text Detection}
% Since the advent of the Transformer Neural architecture \cite{vaswani2017attention}, the field of Natural Language Generation (NLG) has experienced massive improvements \cite{zhao2023survey}.
Large Language Models (LLMs) such as GPT-3.5, GPT-4 \cite{openai2023gpt4}, 
LLaMA \cite{touvron2023llama}, Falcon \cite{refinedweb}, 
have the capacity to generate human-like-quality texts, which can be easily construed as human-written \cite{sadasivan2023can,chakraborty2023possibilities, zhao2023survey}. 
However, while such LLMs are remarkable, it, therefore, makes them susceptible to malicious use. These include the generation of toxic and harmful content, like misinformation and terrorism recruitment \cite{shevlane2023model,zellers2019defending,uchendu2021turingbench}. Due to such potential for misuse, we must develop techniques to distinguish human-written texts from LLM-generated ones to mitigate these risks. 

To mitigate this potential for misuse of LLMs, researchers have developed several types of 
automatic detectors.
%\cite{su2023detectllm, uchendu2023attribution,jawahar2020automatic,guerrero2022synthetic,crothers2022machine}. 
These techniques include supervised \cite{uchendu2021turingbench,zellers2019defending,uchendu2020authorship,zhong2020neural,kushnareva2021artificial,liu2022coco}
and 
unsupervised approaches \cite{gehrmann2019gltr,mitchell2023detectgpt,galle2021unsupervised,he2023mgtbench,su2023detectllm}. 
These supervised approaches tend to be stylometric-, deep learning- and ensemble-based models while most unsupervised approaches are statistical-based detectors \cite{uchendu2023attribution, yang2023survey}. 

More recently, due to the increased ubiquity of LLMs, we need more interpretable, and less 
deep learning-based models. Deep learning models have been shown to be the most susceptible to adversarial perturbations than others \cite{pu2022deepfake}.
To that end, we propose a supervised statistical-based technique, that calculates UID-based features of a given text and uses a classical machine learning model to automatically decide thresholds.

% \begin{table*}[]
%     \centering
%     \footnotesize
% \resizebox{15.5cm}{!}{
% \renewcommand{\arraystretch}{1.25}
%         \begin{tabular}{lccccccc}
%             \hline
%             \multicolumn{1}{c}{\textbf{Span Length (N)}} & \textbf{N=20} & \textbf{N=0} & \textbf{N=4} & \textbf{N=10} & \textbf{N=15} & \textbf{N=20} & \textbf{N=30} \\
%             \hline
%             & \textbf{Random UID} & \textbf{UID Features} & \multicolumn{5}{c}{\textbf{Min + Max UID Spans}} \\
%             & \textbf{Spans} & \textbf{Only} & & & & \\
%             \hline
%             \textbf{TuringBench} & 0.62 & 0.65 & 0.82 & 0.84 & 0.87 & \textbf{0.88} & 0.86 \\
%             \textbf{InTheWild} & 0.72 & 0.75 & 0.79 & 0.83 & 0.86 & \textbf{0.88} & 0.87 \\
%             \hline
%         \end{tabular}}
%     \caption{Max. \& Min. UID spans ablation study: Setting a span length of N=20 tokens maximized performance across large-scale datasets (N>30 subsequently lower and eventually consistent performance). It can be seen that our min/max features tremendously impact performance against a random sample/no span features at all.}
% \label{tab:n_ablation}
% \end{table*}

\section{Our Proposal: {\gptwho}}
\begin{figure}[ht]
\centering
    \includegraphics[width=\columnwidth]{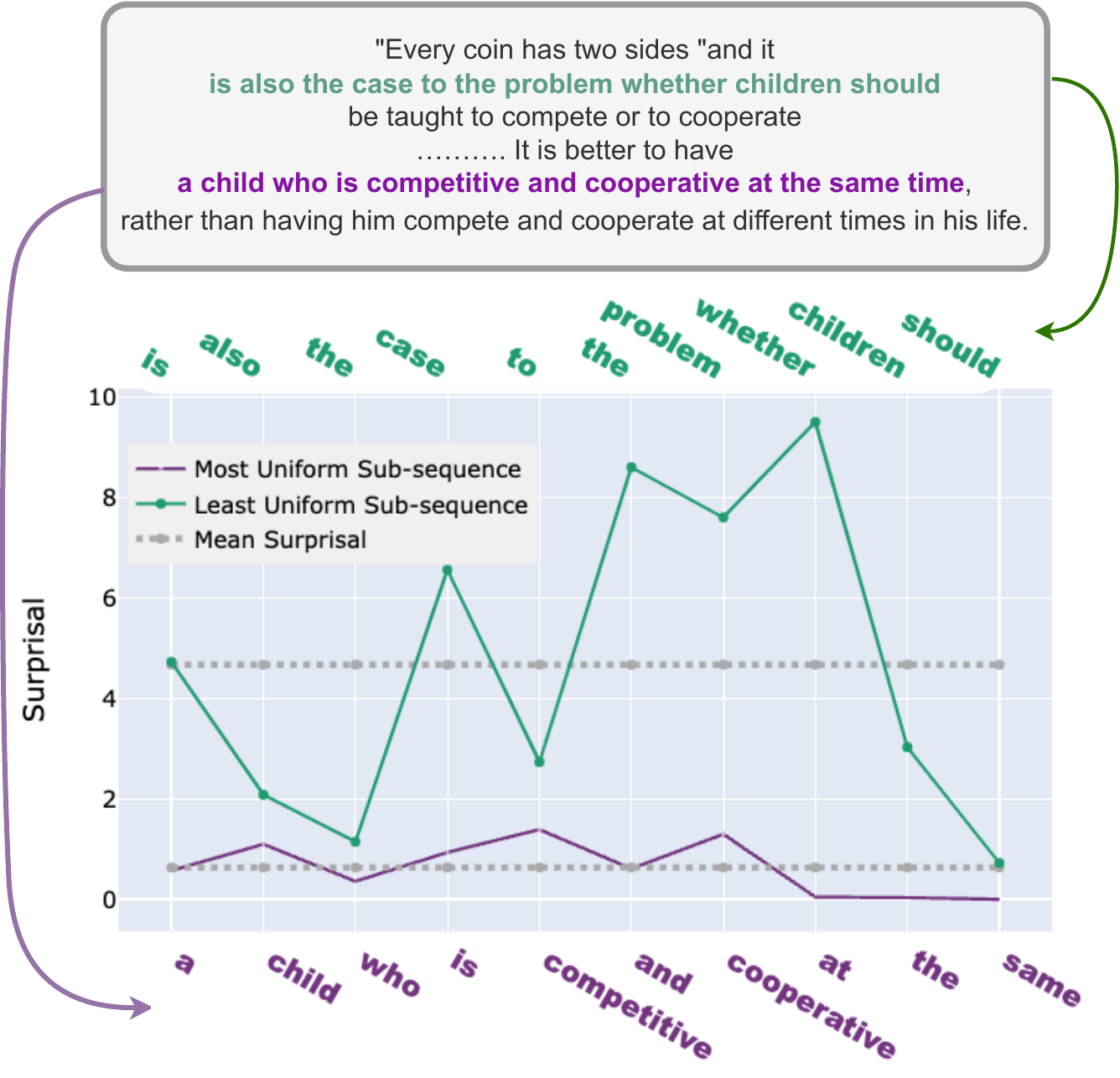}
  \caption{An example of UID span feature extraction that selects the most uniform and non-uniform segments from the token surprisal sequence. As can be seen in this example, two texts that read well can have very different underlying information density distributions in a given context. UID features capture these hidden statistical distinctions that are not apparent in their textual form.  }
  \label{fig:interpret}
\end{figure}

We propose a psycholinguistically-motivated statistical-based machine-generated text detector {\gptwho} that uses a \textsf{\textbf{GPT}}-based language model to predict \textsf{\textbf{who}} the author of an article is. {\gptwho} works by exploiting a densely information-rich feature space motivated by the UID principle. UID-based representations are sensitive to intricate ``fluctuations'' as well as ``smoothness'' in the text. Specifically, operationalizations of UID are aimed at capturing the evenness or smoothness of the distribution of surprisal per linguistic unit (tokens, words), as stated by the UID principle. For example, in Figure \ref{fig:interpret}, we show sequences of tokens that correspond to the highest and lowest UID score spans within an article. Here, the differences between the two segments of texts might not be obvious at the linguistic level to a reader, but when mapped to their surprisal distributions, the two segments have noticeably distinct surprisal spreads as can be seen by the peaks and troughs i.e. variance of token surprisals along the y-axis about the mean (dotted line). Most approximations of this notion of ``smoothness'' of information spread and UID, thus, formulate it as the variance of surprisal or as a measure of the difference of surprisals between consecutive linguistic units 
\citep{jain-etal-2018-uniform, meister-etal-2020-beam, regularizer, venkatraman2023decoding}. 

\begin{figure*}[ht!]
  \centering
  \includegraphics[width=0.8\textwidth]{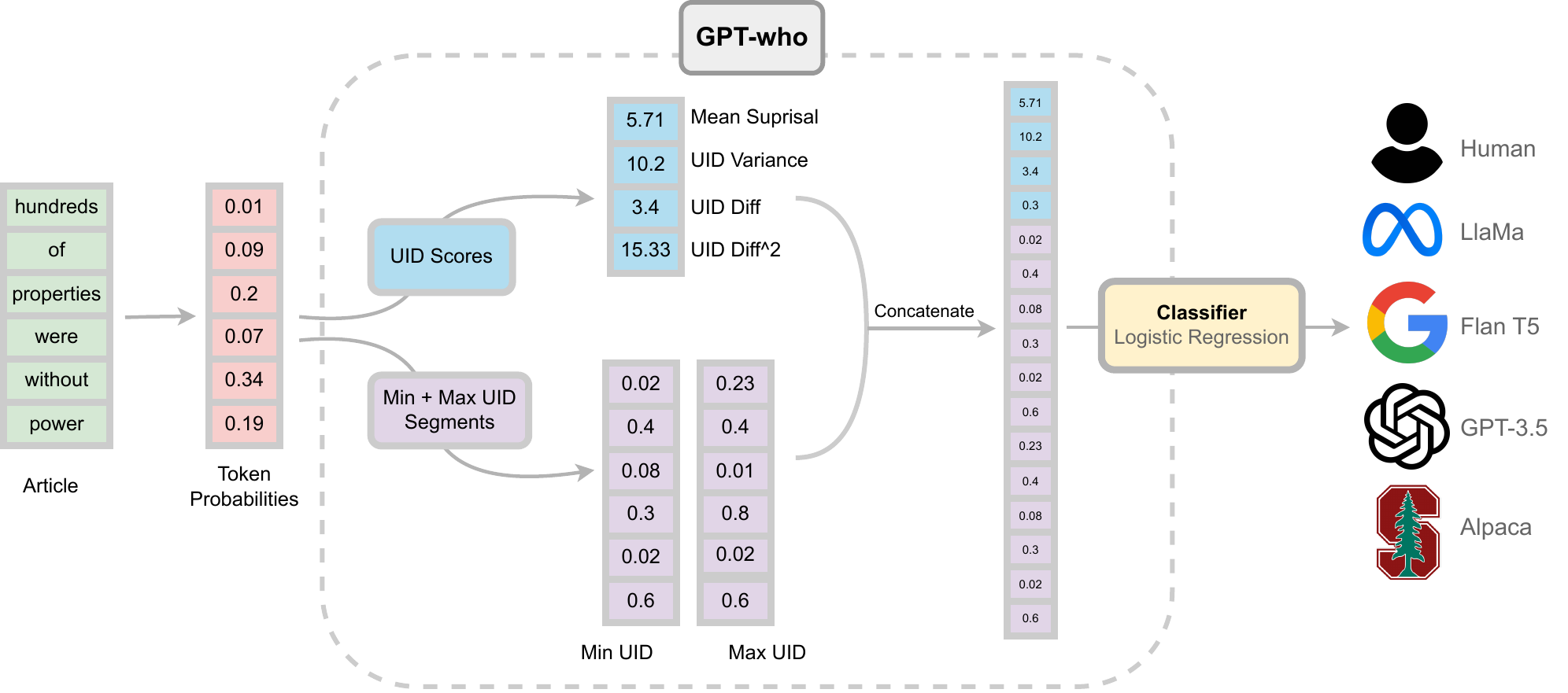}
  \caption{{\gptwho} uses token probabilities of articles to extract UID-based features. A classifier then learns to map UID features to different authors, and identify the author of a new unseen article.}
  \label{fig:gptwho}
\end{figure*}

In measuring the distribution of surprisal of tokens, UID-based features can capture and amplify subtle information distribution patterns that constitute distinct information profiles of authors. Using just an off-the-shelf language model to calculate UID-based features, {\gptwho} learns to predict authorship by means of a simple classifier using UID representations. In addition, as these features can be directly mapped to their linguistic token equivalents, {\gptwho} offers a more interpretable representation of its detection behavior, unlike current black-box statistical detectors, as illustrated in Figure \ref{fig:interpret}. 
The use of a psycholinguistically motivated representation also enables us to better interpret the resulting representation space. It can capture surprisal distributions indicative of and commonly occurring in human-written or machine-generated text. 
{\gptwho} is one of the first text detectors that focus on informing a simple classifier with theoretically motivated and intuitive features, as it only requires a fixed-length UID-based representation of length 44 and learns to predict authorship based on just these features, without the need for the full text or any LM fine-tuning in the process (See {\gptwho}'s complete pipeline in Figure \ref{fig:gptwho}).

%Our results indicate that UID is all you need

\subsection{UID-based features} \label{sec:features}

We use the 3 most widely used measures of UID scores as defined in previous works \citep{jain-etal-2018-uniform, meister-etal-2020-beam, regularizer, venkatraman2023decoding} as follows:  We first obtain the conditional probability \textit{p} of each token ($y_{t}$) in an article using a pre-trained LM (GPT2-XL). The surprisal ($u$) of a token $y_{t}$ is, 
\newcommand\eqdef{\ensuremath{\stackrel{\rm def}{=}}} % Equal by definition
\begin{align} 
    u(y_{t}) = - \log  (p(y | y<t)),
\end{align}    
for $t \geq 1$ where $y_{0} = <BOS>$, and $t$ = time step. 

The lower the probability of a token, the higher its surprisal and vice-versa. Thus, surprisal indicates how unexpected a token is in a given context. 
\begin{enumerate}
    
\item \textbf{Mean Surprisal ($\mu$)} of an article (\textit{y}) of length {|y|} in number of tokens is defined as follows:
\begin{align}
    \mu(y) = \frac{1}{|y|} \sum_{t}(u(y_{t}))
\end{align} 

\item \textbf{UID ($Variance$)} score or \textbf{global} UID score of an article (\textit{y}) is calculated as the normalized variance of the surprisal:
\begin{align}
    \mathrm{UID}(y) =  \frac{1}{|y|} \sum_{t}(u(y_{t}) -  \mu )^{2}  
\end{align}
From this formulation, a perfectly uniform article would have the same surprisal at every token and hence $0$ UID (variance) score. 

\item \textbf{UID ($Difference$)} score or \textbf{local} UID score of an article (\textit{y}) is calculated as the average of the difference in surprisals of every two consecutive tokens $\mu(y_{t-1})$ and $\mu(y_{t})$ :
\begin{align}
 \mathrm{UID}(y)  = \frac{1}{{|y|}-1} \sum_{t=2}^{|y|} abs(\mu\left(y_t\right)-\mu\left(y_{t-1}\right))
\end{align}
\item \textbf{UID ($Difference^2$)} score is defined as the average of the squared difference in surprisals of every two consecutive tokens $\mu(y_{t-1})$ and $\mu(y_{t})$ :
\begin{align}
 \mathrm{UID}(y)  = \frac{1}{{|y|}-1} \sum_{n=2}^{|y|} (\mu\left(y_t\right)-\mu\left(y_{t-1}\right))^2 
\end{align}

From this formulation, both local measures of UID capture any sudden bursts of unevenness in how information is dispersed in consecutive tokens of the articles.
\end{enumerate}

\begin{table*}[ht!]
 \centering
 \small
    \resizebox{12.5cm}{!}{
        \renewcommand{\arraystretch}{1.25}
        \begin{tabular}{lccccccc}
            \toprule
            & \multirow{2}{*}{Random} & \multirow{2}{*}{No Spans} & \multicolumn{5}{c}{Span Length (N) of Min/Max UID spans} \\
            \cmidrule{4-8}
            Human v. & UID spans & & N=4 & N=10 & N=15 & N=20 & N=30 \\
            \midrule
            GPT-1 & 0.75 & 0.76 & 0.99 & 0.99 & 0.98 & \textbf{1.00} & \underline{0.99} \\
            GPT-2\_small & 0.62 & 0.64 & 0.75 & 0.82 & \textbf{0.88} & \textbf{0.88} & \underline{0.85} \\
            GPT-2\_medium & 0.63 & 0.63 & 0.73 & 0.80 & \textbf{0.88} & \underline{0.87} & 0.84 \\
            GPT-2\_large & 0.65 & 0.62 & 0.73 & 0.79 & \textbf{0.88} & \textbf{0.88} & \underline{0.83} \\
            GPT-2\_xl & 0.65 & 0.61 & 0.72 & 0.80 & \underline{0.88} & \textbf{0.89} & 0.85 \\
            GPT-2\_PyTorch & 0.55 & 0.64 & 0.83 & 0.84 & \textbf{0.87} & 0.85 & \underline{0.86} \\
            GPT-3 & 0.63 & 0.69 & 0.71 & 0.73 & \underline{0.77} & \textbf{0.84} & 0.74 \\
            GROVER\_base & 0.63 & 0.65 & 0.76 & 0.77 & \underline{0.79} & \textbf{0.81} & 0.78 \\
            GROVER\_large & 0.59 & 0.60 & 0.71 & 0.71 & \underline{0.73} & \textbf{0.75} & 0.72 \\
            GROVER\_mega & 0.55 & 0.56 & 0.67 & 0.67 & \underline{0.68} & \textbf{0.72} & 0.67 \\
            CTRL & 0.79 & 0.83 & \textbf{0.99} & \underline{0.98} & \underline{0.98} & \textbf{0.99} & \underline{0.98} \\
            XLM & 0.62 & 0.69 & \underline{0.96} & \underline{0.96} & \underline{0.96} & \textbf{0.99} & \underline{0.96} \\
            XLNET\_base & 0.62 & 0.71 & 0.95 & 0.97 & \underline{0.98} & \underline{0.98} & \textbf{0.99} \\
            XLNET\_large & 0.49 & 0.70 &\underline{0.99} & \underline{0.99} & \underline{0.99} & \textbf{1.00} & \underline{0.99} \\
            FAIR\_wmt19 & 0.54 & 0.57 & 0.74 & 0.75 & \textbf{0.78} & 0.74 & \underline{0.76} \\
            Fair\_wmt20 & 0.62 & 0.63 & 0.72 & 0.75 & 0.88 & \textbf{1.00} & \underline{0.89} \\
            TRANSFO\_XL & 0.70 & 0.70 & 0.79 & 0.80 &\underline{0.83} & 0.79 &\textbf{0.84} \\
            PPLM\_distil & 0.57 & 0.62 & 0.92 & 0.91 & \underline{0.93} &\textbf{0.95} & \underline{0.93} \\
            PPLM\_gpt2 & 0.54 & 0.58 & 0.88 & 0.88 & \textbf{0.90} & \underline{0.89} & 0.88 \\
            \midrule
            TuringBench (Avg F1) & 0.62 & 0.65 & 0.82 & 0.84 & \underline{0.87} & \textbf{0.88} & 0.86 \\
            \midrule
            InTheWild (Avg F1) & 0.72 & 0.75 & 0.79 & 0.83 & 0.86 & \textbf{0.88} & \underline{0.87} \\
            \bottomrule
        \end{tabular}}
\caption{Max. \& Min. UID spans ablation study: Setting a span length of N=20 tokens maximized performance (F1 score) across large-scale datasets (N>30 leads to subsequently lower and eventually consistent performance). It can be seen that our min/max features tremendously impact performance against randomly sampled or no span features at all.}
\label{tab:n_ablation_full}
\end{table*}

\textbf{Maximum and minimum UID spans} 
In addition to previously used approximations of UID, we also craft a new set of features using the most and least uniform segments of an article. Our intuition for this feature is to focus on the extremities of the UID distribution in an article, as the most and least uniform spans would be the most expressive and distinct sequences from a UID perspective. All other spans or segments in an article necessarily lie in between these two extremities. Thus taking account of these two spans would ensure coverage of the whole range of surprisal fluctuations within an article. Thus, for each article, we calculate UID (variance) scores for all spans of consecutive tokens of a fixed length using a sliding window approach. We tuned this window size and found that a window size of $20$ tokens per span sufficiently represented an article's UID range. We also experimented with randomly drawn and re-ordered spans and found that random features did not contribute to task performance (see Table \ref{tab:n_ablation_full} for ablation study results). We use the surprisal values corresponding to the highest and lowest UID scoring span as additional features and obtain fixed length UID features of length 44 for each article. 

\section{Empirical Validation}

\begin{figure*}[ht]
  \centering
  \includegraphics[width=0.83\textwidth]{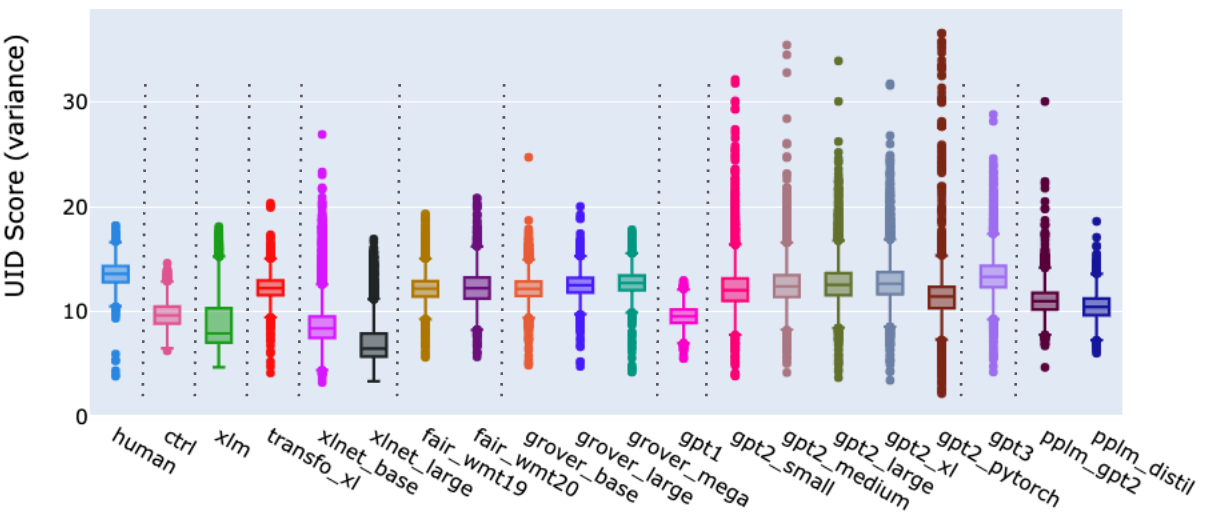}
  \caption{Distribution of UID Scores of 20 authors from the TuringBench dataset grouped (dotted line) by architecture type. LMs that share architectures tend to distribute UID scores similarly.}
  \label{fig:turingbench_uid}
\end{figure*}

We use \citet{meister2021revisiting}'s implementation of UID-based scores\footnote{\url{https://github.com/rycolab/revisiting-uid/tree/main}} and use the publicly available off-the-shelf pre-trained GPT2-XL language model\footnote{\url{https://huggingface.co/gpt2-xl}} to obtain conditional probabilities. For all our experiments, we calculate the UID features for the publically released train and test splits of all datasets of each of the 4 benchmarks as they were released by the dataset developers. We train a logistic regression model\footnote{\url{https://scikit-learn.org/stable/}} using these features on the train splits and report performance on the test splits. We averaged performance over 3 different random seeds and set the number of maximum iterations hyperparameter to 10k after testing a set of values. We replicate all the original evaluation settings and metrics for each of the datasets (except one setting from the ArguGPT \cite{liu2023argugpt} dataset that required access to unreleased human evaluation data). We do this to be able to directly compare the performance of {\gptwho} with current state-of-the-art detection methods reported so far.  

\subsection{Datasets} \label{sec:datasets}
To test the applicability of {\gptwho} across text detection tasks, we run all experiments across 4 large-scale and recent datasets that span over 15 domains and 35 recent LMs. 

\paragraph{TuringBench Benchmark 
\cite{uchendu2021turingbench}} dataset is the largest multi-class authorship attribution dataset that contains over 168k news articles generated by 19 neural text generators using 10K prompts from CNN and the Washington Post.

\paragraph{GPABenchmark \cite{liu2023check}} or \underline{G}PT Cor\underline{p}us for \underline{A}cademia is a multi-domain (Computer Science (CS), Humanities and Social Sciences (HSS) and Physics (PHX)) academic articles dataset aimed at helping detection of LLM use or misuse in academic writing. It contains 150k human and 450k ChatGPT-generated articles for 3 task settings (completion, writing, and polishing). 

\paragraph{ArguGPT \cite{liu2023argugpt}}
is a prompt-balanced dataset of argumentative essays containing over 4k human-written essays and 4k articles generated by 7 recent LLMs (including many variants of ChatGPT) using prompts from English datasets such as TOEFL11 \cite{blanchard2013toefl11} and WECCL \cite{wen2005spoken} datasets.

\paragraph{``InTheWild'' Deepfake Text Detection in the Wild \cite{li2023deepfake}} dataset is, to our knowledge, the largest text detection dataset consisting of over 447k human-written and machine-generated texts from 10 tasks such as story generation, news article writing, and academic writing. They use 27 recent LLMs such as GPT-3.5, FLAN-T5, and LLaMA. We refer to this dataset as the \textbf{``InTheWild''} dataset going forward for brevity. 

\subsection{Baselines \& Detectors} We compare our proposed method against the following: DetectGPT \footnote{\url{https://github.com/eric-mitchell/detect-gpt}} \cite{mitchell2023detectgpt}, GLTR\footnote{\url{https://github.com/HendrikStrobelt/detecting-fake-text}} \cite{gehrmann2019gltr}, an open-source implementation\footnote{\url{https://github.com/BurhanUlTayyab/GPTZero}} of GPTZero \cite{tian2023gptzero}, ZeroGPT \cite{zerogpt}, OpenAI's detector \cite{openaidetector}, \citet{li2023deepfake}'s LongFormer-based detector\footnote{\url{https://github.com/yafuly/DeepfakeTextDetect}} tuned for the InTheWild benchmark (we refer to this method as \textbf{``ITW''}), a stylometric detector\footnote{\url{https://github.com/shaoormunir/writeprints}} \cite{abbasi2008writeprints} and fine-tuned BERT\footnote{\url{https://huggingface.co/docs/transformers/training}} \cite{bert}. We are unable to report results for exhaustively all methods across all datasets due to inherent inapplicability in certain task settings. For example, most SOTA text detectors cannot be applied to the ArguGPT dataset as it only contains text written by multiple machines, while most text detectors are designed to differentiate between human-written and machine-generated texts. Beyond such limitations, we have utilized all applicable methods for 4 benchmark datasets.

\subsection{UID Signatures of Authors} 
Given that humans tend to optimize UID, we study if different models spread surprisal in ways that are distinguishable from each other and human-written text and if we can observe unique UID signatures of different LM families. To this end, we plot the UID score distributions of different text generators across (see Figures \ref{fig:turingbench_uid}, \ref{fig:model_diffs}, and \ref{fig:task_diffs}). We observe that, generally, the UID scores of human-written text have a higher mean and larger standard deviation than most machine-written text across writing task types, domains, and datasets. This implies that human-written text tends to be more non-uniform and diverse in comparison to machine-generated text. Hence, machines seem to be spreading information more evenly or smoothly than humans who are more likely to have fluctuations in their surprisal distributions. Going a step further, if we compare models to other models, we see that models that belong to the same LM family by architecture tend to follow similar UID distribution. For example, in Figure \ref{fig:turingbench_uid}, the dotted lines separate LMs by their architecture type and it can be seen, for example, that all GPT-2 based models have similar UID distributions, all Grover-based models have similarities, but these groups are distinct from each other. This indicates that UID-based features can capture differences in text generated by different LM families. To our knowledge, this is the first large-scale UID-based analysis of recent machine and human-generated text across writing tasks and 
domains.  \begin{table*}
    \centering
    \small
    \renewcommand{\arraystretch}{1.25}
    \begin{tabular}{llccccccc}
        \toprule
        Task Type & Domain & GPTZero & ZeroGPT & OpenAI Detector & DetectGPT & BERT & ITW & {\gptwho} \\
        \midrule

        & CS & 0.30 & 0.67 & 0.81  & 0.58  & \textbf{0.99} & \underline{0.98} & \textbf{0.99} \\
        & PHX & 0.25 & 0.68  & 0.70  & 0.54  &  \textbf{0.99} & \underline{0.98} & \underline{0.98} \\ 
        \multirow{-3}{*}{Task 1} & HSS & 0.72 & 0.92  & 0.63  & 0.57  & \textbf{0.99} & 0.96 & \underline{0.98}\\
        \midrule
         & CS& 0.17  & 0.25  & 0.64 & 0.16 & \textbf{0.99}  & 0.81 & \underline{0.84} \\
         & PHX& 0.06 & 0.10 & 0.24 & 0.17 & \textbf{0.96}& 0.76  & \underline{0.90} \\
        \multirow{-3}{*}{Task 2} & HSS & 0.44   & 0.62 & 0.27& 0.20 & \textbf{0.97}  & 0.29 & \underline{0.80}\\  
        \midrule
         & CS  & 0.02 & 0.03  & 0.06  & 0.03  & \textbf{0.97} & 0.38 & \underline{0.63} \\
         & PHX  & 0.02 & 0.03  & 0.04  & 0.05  & \textbf{0.97} & 0.31  & \underline{0.75}\\
        \multirow{-3}{*}{Task 3} & HSS  & 0.20 & 0.25  & 0.06 & 0.06  & \textbf{0.99}  & 0.08  & \underline{0.62}\\ 
        \midrule
        Average F1 & & 0.24 & 0.40 & 0.38 & 0.26 & \textbf{0.98} & 0.62 & \underline{0.83} \\
        \bottomrule
    \end{tabular}
    \caption{Test Set Performance (F1 Scores) of different machine text detectors on the GPA Benchmark. Best performance are in bold, and second best underlined. } 
    \label{tab:gpabench_result}
\end{table*}
% * indicate results reported in \citet{liu2023check}.
\begin{table*}
    \centering
    \small
    \renewcommand{\arraystretch}{1.25}
    \begin{tabular}{lccccccccc}
        \toprule
       Human v.       & GROVER & GTLR & GPTZero & DetectGPT & RoBERTa & BERT & ITW & Stylometry & {\gptwho} \\
       \midrule 
GPT-1           & 0.58                       & 0.47                      & 0.47                        & 0.51                          & 0.98                        & 0.95                     & 0.92                    & \underline{0.99}                     & \textbf{1.00}               \\
GPT-2\_small    & 0.57                       & 0.51                      & 0.51                        & 0.51                          & 0.71                        & \underline{0.75}               & 0.47                    & \underline{0.75}                     & \textbf{0.88}               \\
GPT-2\_medium   & 0.56                       & 0.49                      & 0.50                        & 0.52                          & \underline{0.75}                  & 0.65                     & 0.47                    & 0.72                           & \textbf{0.87}               \\
GPT-2\_large    & 0.55                       & 0.46                      & 0.49                        & 0.51                          & \underline{0.79}                  & 0.73                     & 0.46                    & 0.72                           & \textbf{0.88}               \\
GPT-2\_xl       & 0.55                       & 0.45                      & 0.51                        & 0.51                          & 0.78                        & \underline{0.79}               & 0.45                    & 0.73                           & \textbf{0.89}               \\
GPT-2\_PyTorch  & 0.57                       & 0.72                      & 0.50                        & 0.52                          & 0.84                        & \textbf{0.99}            & 0.47                    & 0.83                           & \underline{0.85}                  \\
GPT-3           & 0.57                       & 0.35                      & 0.47                        & 0.52                          & 0.52                        & \underline{0.79}               & 0.48                    & 0.72                           & \textbf{0.84}               \\
GROVER\_base    & 0.58                       & 0.39                      & 0.52                        & 0.51                          & \textbf{0.99}               & \underline{0.98}               & 0.49                    & 0.76                           & 0.81                        \\
GROVER\_large   & 0.54                       & 0.41                      & 0.47                        & 0.52                          & \textbf{0.99}               & \underline{0.98}               & 0.52                    & 0.71                           & 0.75                        \\
GROVER\_mega    & 0.51                       & 0.42                      & 0.42                        & 0.51                          & \underline{0.94}                  & \textbf{0.97}            & 0.53                    & 0.68                           & 0.72                        \\
CTRL            & 0.49                       & 0.88                      & 0.67                        & 0.67                          & \textbf{1.00}               & \textbf{1.00}            & 0.91                    & \underline{0.99}                     & \underline{0.99}                  \\
XLM             & 0.50                       & 0.89                      & 0.67                        & 0.67                          & 0.58                        & \textbf{1.00}            & 0.92                    & 0.96                           & \underline{0.99}                  \\
XLNET\_base     & 0.58                       & 0.75                      & 0.51                        & 0.67                          & 0.79                        & \textbf{0.99}            & 0.84                    & 0.95                           & \underline{0.98}                  \\
XLNET\_large    & 0.58                       & 0.88                      & 0.67                        & 0.52                          & \textbf{1.00}               & \textbf{1.00}            & \underline{0.93}              & \textbf{1.00}                  & \textbf{1.00}               \\
FAIR\_wmt19     & 0.56                       & 0.56                      & 0.56                        & 0.51                          & \underline{0.84}                  & \textbf{0.93}            & 0.49                    & 0.74                           & 0.74                        \\
Fair\_wmt20     & 0.58                       & 0.49                      & 0.50                        & 0.51                          & 0.45                        & 0.47                     & 0.47                    & \underline{0.73}                     & \textbf{1.00}               \\
TRANSFO\_XL & 0.58                       & 0.35                      & 0.49                        & 0.52                          & \underline{0.96}                  & \textbf{0.97}            & 0.81                    & 0.79                           & 0.79                        \\
PPLM\_distil    & 0.59                       & 0.64                      & 0.52                        & 0.67                          & 0.90                        & 0.88                     & 0.51                    & \underline{0.92}                     & \textbf{0.95}               \\
PPLM\_gpt2      & 0.58                       & 0.68                      & 0.51                        & 0.51                          & \textbf{0.90}               & \underline{0.89}               & 0.49                    & 0.88                           & \underline{0.89}                  \\
\midrule
Average F1      & 0.56                       & 0.57                      & 0.52                        & 0.55                          & \textbf{0.88}               & 0.61    & \textbf{0.88}           & \underline{0.82}                     & \textbf{0.88}     \\
    \bottomrule
    \end{tabular}
    \caption{Test Set Performance (F1 score) for TuringBench dataset. Overall, {\gptwho} outperforms both statistical and supervised detectors, and is at part with BERT.}
    \label{tab:turing_bench_result}
\end{table*}

\begin{table*}
     \centering
    \small
    \renewcommand{\arraystretch}{1.25}
    \begin{tabular}{llcccccc}
        \toprule
        Detection Setting & Testbed Type & GPTZero & GLTR & DetectGPT & BERT & ITW & {\gptwho} \\
        \midrule
        % \multicolumn{}{c}{In-distribution Detection} \\
        \multirow{4}{*}{In-distribution} & Domain-specific Model-specific & 0.65 & 0.94 & 0.92 & \textbf{0.98} & \underline{0.97} & 0.93 \\
         & Cross-domains Model-specific & 0.63 & 0.84 & 0.6 & \textbf{0.98} & \underline{0.97} & 0.88 \\
        & Domain-specific Cross-models & 0.57 & 0.8 & 0.57 & 0.49 & \textbf{0.87} & \underline{0.86} \\
        & Cross-domains Cross-models & 0.57 & 0.74 & 0.57 & 0.49 & \underline{0.78} & \textbf{0.86} \\
        \midrule
        % \multicolumn{7}{c}{Out-of-distribution Detection} \\
        \multirow{2}{*}{Out-of-distribution}& Unseen Models & 0.58 & 0.65 & 0.6 & \textbf{0.84} & \underline{0.79} & 0.74 \\
        &Unseen Domains & 0.57 & 0.72 & 0.57 & 0.68 & \textbf{0.8} & \underline{0.77} \\
        \midrule
        & Average F1 & 0.60 & 0.78 & 0.64 & 0.74 & \textbf{0.86} & \underline{0.84} \\
        \bottomrule
    \end{tabular}
    \caption{Test Set Performance (F1 score) for InTheWild dataset. ITW refers to the LongFormer-based detector trained by Li et al. (2023) specifically for this benchmark.}
    \label{tab:itw_result}
\end{table*}

\subsection{Machine Text Detection Performance}
Overall, {\gptwho} outperforms other statistical-based detectors and is at par with transformers-based fine-tuned methods for 2 out of 4 benchmarks. For \textbf{GPABenchmark} (Table \ref{tab:gpabench_result}), across all task types and domains, {\gptwho} outperforms GPTZero, ZeroGPT, DetectGPT and, OpenAI's detector by over \textbf{40\%}. The machine-generated texts for this task are from 7 very recent and highly sophisticated LLMs (including GPT3.5, GPT3 variants), making the detection of machine-generated text a much more challenging task on which {\gptwho} outperforms other detectors. 

For \textbf{TuringBench} (Table \ref{tab:turing_bench_result}), {\gptwho} significantly outperforms GLTR by \textbf{0.32 F1} points, and at par with BERT fine-tuned for the task. The \textbf{InTheWild} dataset contains 6 testbeds with varying levels of detection difficulties, such as out-of-domain, out-of-distribution, and unseen-task test sets. We used all 6 testbeds to analyze the performance of {\gptwho} in detecting machine-generated texts across increasing levels of `wildness' and find that overall, {\gptwho} outperforms all other methods except the one specifically tuned to the task (ITW) across all testbeds. More importantly, {\gptwho} performs well even for the most challenging or `wildest' testbed settings of unseen model and unseen domain distributions (see Table \ref{tab:itw_result}).

\begin{table}
    \centering
    \renewcommand{\arraystretch}{1.25}
    \setlength{\tabcolsep}{2pt} % Adjust column separation as needed
    \small
    \begin{tabular}{lcccc}
        \toprule
        Author           & Experts* & Stylometry & BERT & {\gptwho} \\
        \midrule
        text-babbage-001 & 0.47     & 0.45       & \underline{0.84} & \textbf{0.85}    \\
        text-curie-001   & 0.47     & 0.45       & \underline{0.83} & \textbf{0.84}    \\
        text-davinci-003 & 0.66     & 0.59       & \textbf{0.95} & \underline{0.77}    \\
        gpt-3.5-turbo    & 0.63     & 0.69       & \textbf{0.96} & \underline{0.84}    \\
        gpt2-xl          & 0.37     & 0.49       & \textbf{0.95} & \underline{0.91}    \\
        \midrule
        Average F1       & 0.52     & 0.53       & \textbf{0.91} & \underline{0.84}    \\
        \bottomrule
    \end{tabular}
    \caption{Test Set Performance (F1 score) for ArguGPT dataset.\textsuperscript{*} denotes results reported in \citet{liu2023argugpt}.}
    \label{tab:argugpt_result}
\end{table}
For the \textbf{ArguGPT} dataset (Table \ref{tab:argugpt_result}), we find that {\gptwho} outperforms human experts and stylometry in predicting authorship by \textbf{0.31 F1} points, but is outperformed by fine-tuned BERT. Although unable to perform as well as BERT, {\gptwho} is one of the only statistical-based detectors that can handle distinctions between machine-only texts. We were unable to evaluate other detectors as their human-generated texts were not publicly released, and they only work in human v/s machine settings. 

{\gptwho} is a statistical-based approach that outperforms other statistical-based approaches but is unsurprisingly outperformed by fine-tuned methods in 2 out of 4 benchmarks. In the case of statistical-based methods, it is typically very hard to come close to fine-tuned performance as such methods rely only on derived properties of the text and do not utilize the full raw text in training as is the case in the latter \cite{jawahar2020automatic}. Despite this, {\gptwho} can exceed fine-tuned LM performance by 10\% for 2 benchmarks.

\subsection{Running Time}
We measured the time taken for the one-time training or fine-tuning and inference for 6 testbeds from the InTheWild Dataset (the largest of all our benchmarks). We compare the average running times of DetectGPT, BERT, {\gptwho} and a stylometric detector in Table \ref{tab:running_times} and find that {\gptwho} is the fastest as it eliminates the need for any LM fine-tuning and makes a single inference call per text sample. Other methods require either LM fine-tuning or multiple inference calls (for example,  DetectGPT). This computational load is greater than a single forward inference pass through one LM (GPT2) followed by logistic regression which is what {\gptwho} requires. 

\begin{table}[htbp]
    \centering
    \begin{tabular}{lcc}
        \toprule
        \textbf{Method} & \textbf{One-Time Training} & \textbf{Inference} \\
        \midrule
        DetectGPT & \textgreater{}10 hours & 60 sec \\
        BERT & $\sim$1.5 hours & 2 sec \\
        Stylometry & $\sim$1.5 hours & 2 sec \\
        {\gptwho} & 20 min & 0.8 sec \\
        \bottomrule
    \end{tabular}
    \caption{Average Running time over 6 testbeds from the InTheWild dataset.}
    \label{tab:running_times}
\end{table}

\section{Discussion}
We turn to the UID principle, which states that \textit{humans prefer to spread information evenly in language}, to automatically extract features that measure the spread and flow of information content or surprisal in texts. Our UID-based features are formulated to capture how surprisal is distributed in an article as they measure the local and global variance, mean, and most uniform and non-uniform segments of a text. This rich and succinct representation space drives the predictive capability of our proposed detector and the interpretability of its representations. Analysis of this feature space reveals that \textbf{human-written text tends to be more non-uniform in comparison to machine-generated text}. Hence, machines seem to be spreading information more evenly or smoothly than humans who are more likely to have fluctuations in their surprisal distributions. However, this finding does not imply that humans are not producing uniform text. It is important to note that our work cannot provide support for or refute the UID hypothesis which comes from psycholinguistic studies such as those in Section \ref{sec:UID}. Our work shows that, given our operationalization of UID based on prior works, machine text is relatively more uniform than human-written text. While this might seem contradictory to UID theory, it does not still disprove that humans are uniform in their language production.   

We conjecture that this unexpected finding is because we use GPT-2’s probability distribution to calculate surprisal, which is potentially a poor approximation of the ``human'' probability distribution. A closer-to-human probability distribution might (or might not) show humans to be more uniform than machines, though this determination is not within the scope of this work. It is crucial to note that uniformity is relative, and while machines are more uniform under this operationalization, it would still be true that human text is uniform as per the human’s probability distribution (that we do not have access to and can only approximate using some LM distribution, for example, GPT2-XL in our case). UID theory does not make any predictions on where machine-generated text might lie in the uniform to non-uniform spectrum but only indicates that humans are arranging utterances evenly as per their own language distribution. 

Irrespective of its alignment with what theory suggests, we find that the UID-based features are very useful in distinguishing authors, which is the focus of this work. This is an important consideration and helps disentangle the utility of UID-inspired features from the cognitive plausibility of those feature calculations or UID approximations. 

Thus, this operationalization of UID does not imply that humans are ``less human'' or machines are ``more human'' as it is an approximation of a theory that states that humans are uniform as per their language distribution. It does not have any further implications for machine-generated text and is unable to predict what happens in the case of machines. 

We find that UID-based features can capture differences between text generated by not only humans and models but also capture differences between multiple models and LM families. Our main contribution is a psycholinguistically-inspired domain-agnostic statistical-based machine-generated text detector, {\gptwho}, that:

\begin{itemize}
    \item Outperforms statistical approaches across 4 large-scale benchmark datasets that include texts from over 35 LLMs across more than 10 domains.
    \item Generalizes better to out-of-distribution datasets than SOTA detectors.
    \item Computationally more efficient than other supervised detectors as it does not require the fine-tuning or training of any LLMs.
    \item Interpretable due to its psycholinguistically motivated UID-based feature space. 
\end{itemize} 
While our detector may not significantly outperform fine-tuned transformers-based models, it is essential to highlight its independence from fine-tuning, offering nearly comparable performance at significantly lower computational costs and remains one of the only statistical-based detectors that can operate in multi-author settings beyond the Turing Test. These findings indicate that approaches rooted in psycholinguistic theories that delineate indicators of ``human-like'' language use hold enormous and untapped potential in tackling the fast catapulting and ever-changing LLM landscape. This work has implications for cognitively plausible and explainable solutions to complex challenges arising from ever-growing automated text generators. 

\section{Conclusion}

We propose {\gptwho}, a statistical-based machine-generated text detector that utilizes features inspired by the Uniform Information Density (UID) principle rooted in the observation that humans tend to evenly distribute information or surprisal in language.
We leverage UID-based features, including variance of surprisals and minimally/maximally uniform sub-sequences extracted from the surprisal sequence generated by an off-the-shelf LM. We demonstrate that these features are highly effective in discerning machine-generated text from human-generated text as they capture nuances in how models and humans distribute surprisal in their texts. Our findings have implications for enhanced text authenticity assessment.

\section*{Limitations}
In our pursuit of a comprehensive examination of texts produced by recent large language models, we encountered limitations arising from resource constraints and the availability of publicly accessible datasets. These factors constrained our ability to encompass a more diverse array of models and tasks, including summarization and question-answering. Furthermore, our study did not delve into whether UID-based methods extend their utility beyond detecting machine-generated text to identify potential issues such as misinformation and plagiarism. We acknowledge these constraints as part of our ongoing commitment to refining and expanding our efforts in future research endeavors.

\section*{Ethical Statement}
It is important to note that there are inherent limitations of AI-based tools and automated machine text detectors such as in this work. Acknowledging the fallibility of these detectors, particularly in generating false positives, we note that there is still a crucial need for human oversight and discretion in the usage of such detectors in real-world settings. For example, ethical concerns surrounding over-vigilance in scrutinizing student-written text are an important consideration for striking a balance between the convenience of automated detection and the preservation of academic integrity. By advocating for responsible development and implementation, we hope to contribute to a landscape where ethical considerations guide the integration of automatic text detection systems in educational settings, safeguarding against undue reliance and promoting fairness, equity, and respect for individual expression.

\section*{Acknowledgments}
This work was in part supported by National Science Foundation (NSF) awards \#1820609, \#1950491, and \#2131144.
\bibliography{main}
\clearpage
\appendix

\section{Appendix} 

\subsection{UID Score distributions of authors}
\begin{figure}[ht!]
         \centering
  \begin{subfigure}[b]{0.9\textwidth}
         \centering
    \includegraphics[width=0.9\textwidth]{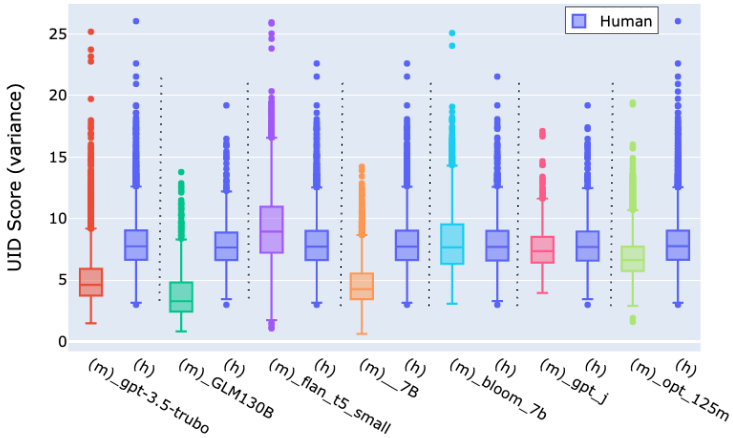}
    \caption{Pairwise comparisons of human and different machine-generated texts for shared tasks: Distribution of UID Scores of 8 authors (7 models + human) from the InTheWild dataset. (m) indicates machine and (h) indicates human written texts. This is followed by the model name along the x-axis labels to indicate the different authors.}
    \label{fig:model_diffs}
  \end{subfigure}
    \vspace{1cm}

  \begin{subfigure}[b]{0.9\textwidth}
         \centering
        \includegraphics[width=0.9\textwidth]{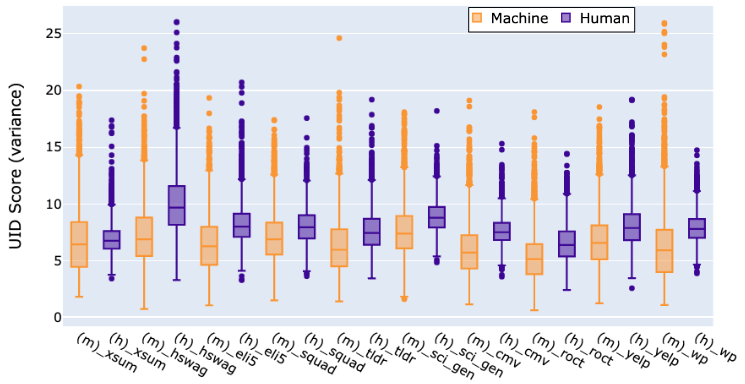}
        \caption{Pairwise comparisons of human and different machine-generated texts for different tasks: Distribution of UID Scores of humans v.s. machines per task type. (m) indicates machine and (h) indicates human written texts. This is followed by the writing task type along the x-axis labels to indicate the different tasks.}
    \label{fig:task_diffs}
    \end{subfigure}
\end{figure}

We see that for most cases, humans have a higher UID (variance) score than machines, as can be seen by the higher means of their scores in the box plots. This holds when comparing human-written texts with multiple machine-generated texts over shared tasks (Figure \ref{fig:model_diffs}), and also when comparing their differences between tasks (Figure \ref{fig:task_diffs}). 
\label{sec:appendix}

\end{document}